\documentclass[letterpaper]{article} 
\usepackage{aaai23}  
\usepackage{times}  
\usepackage{helvet}  
\usepackage{courier}  
\usepackage[hyphens]{url}  
\usepackage{graphicx} 
\usepackage{natbib}  
\usepackage{caption} 
\usepackage{pifont}
\usepackage{subfigure}
\usepackage{multirow}
\usepackage{verbatim}
\usepackage{amsmath}
\usepackage{amsfonts}
\usepackage[ruled,vlined]{algorithm2e}
\usepackage{xcolor,colortbl} 
\title{Domain Decorrelation with Potential Energy Ranking}
\author {
    Sen Pei\textsuperscript{\rm 1,2},
    Jiaxi Sun\textsuperscript{\rm 1,2},
    Richard Yi Da Xu\textsuperscript{\rm 4},
    Shiming Xiang\textsuperscript{\rm 1,2}, and 
    Gaofeng Meng\textsuperscript{\rm 1,2,3}\thanks{Corresponding author.}
}
\affiliations {
    \textsuperscript{\rm 1} NLPR, Institute of Automation, Chinese Academy of Sciences\\
    \textsuperscript{\rm 2} School of Artificial Intelligence, University of Chinese Academy of Sciences\\
    \textsuperscript{\rm 3} CAIR, HK Institute of Science and Innovation, Chinese Academy of Sciences\\
    \textsuperscript{\rm 4} FSC1209, Kowloon Tong Campus, Hong Kong Baptist University\\
    \texttt{\{peisen,sunjiaxi\}2020@ia.ac.cn, xuyida@hkbu.edu.hk, \{smxiang,gfmeng\}@nlpr.ia.ac.cn}
}

\usepackage{bibentry}

\begin{document}

\maketitle

\begin{abstract}
Machine learning systems, especially the methods based on deep learning, enjoy great success in modern computer vision tasks under ideal experimental settings. Generally, these classic deep learning methods are built on the \emph{i.i.d.} assumption, supposing the training and test data are drawn from the same distribution independently and identically. However, the aforementioned \emph{i.i.d.} assumption is, in general, unavailable in the real-world scenarios, and as a result, leads to sharp performance decay of deep learning algorithms. Behind this, domain shift is one of the primary factors to be blamed. In order to tackle this problem, we propose using \textbf{Po}tential \textbf{E}nergy \textbf{R}anking (PoER) to decouple the object feature and the domain feature in given images, promoting the learning of label-discriminative representations while filtering out the irrelevant correlations between the objects and the background. PoER employs the ranking loss in shallow layers to make features with identical category and domain labels close to each other and vice versa. This makes the neural networks aware of both objects and background characteristics, which is vital for generating domain-invariant features. Subsequently, with the stacked convolutional blocks, PoER further uses the contrastive loss to make features within the same categories distribute densely no matter domains, filtering out the domain information progressively for feature alignment. PoER reports superior performance on domain generalization benchmarks, improving the average top-1 accuracy by at least 1.20\% compared to the existing methods. Moreover, we use PoER in the ECCV 2022 NICO Challenge, achieving top place with only a vanilla ResNet-18 and winning the \emph{jury award}. The code has been made publicly available at: \textcolor{magenta}{\texttt{https://github.com/ForeverPs/PoER}}.
\end{abstract}

\section{Introduction}
Deep learning methods have been proved to be increasingly effective in many complex machine learning tasks, such as large-scale image classification, objects detection and image generation, to name a few. Generally, the human-surpassing performance that deep neural networks enjoy is greatly benefited from the \emph{i.i.d.} assumption, supposing the training and the test data are drawn from the same distribution independently and identically. Unfortunately, in open-world scenarios, it is difficult to guarantee that the \emph{i.i.d.} assumption always holds, and as a result, leads to sharp performance drop in the presence of inputs from unseen domains. Formally, the aforementioned problem is termed domain generalization (DG). Given several labeled domains (\emph{a.k.a.} source domains), DG aims to train classifiers only with data from these labeled source domains that can generalize well to any unseen target domains. Different from the closely related domain adaptation (DA) task, DG has no access to the data from target domains while DA can use that for finetuning, namely the adaptation step.

\begin{figure}[t]
\centering
\subfigure[category-wise]{
\begin{minipage}[t]{0.5\columnwidth}
\centering
\includegraphics[width=1.55in]{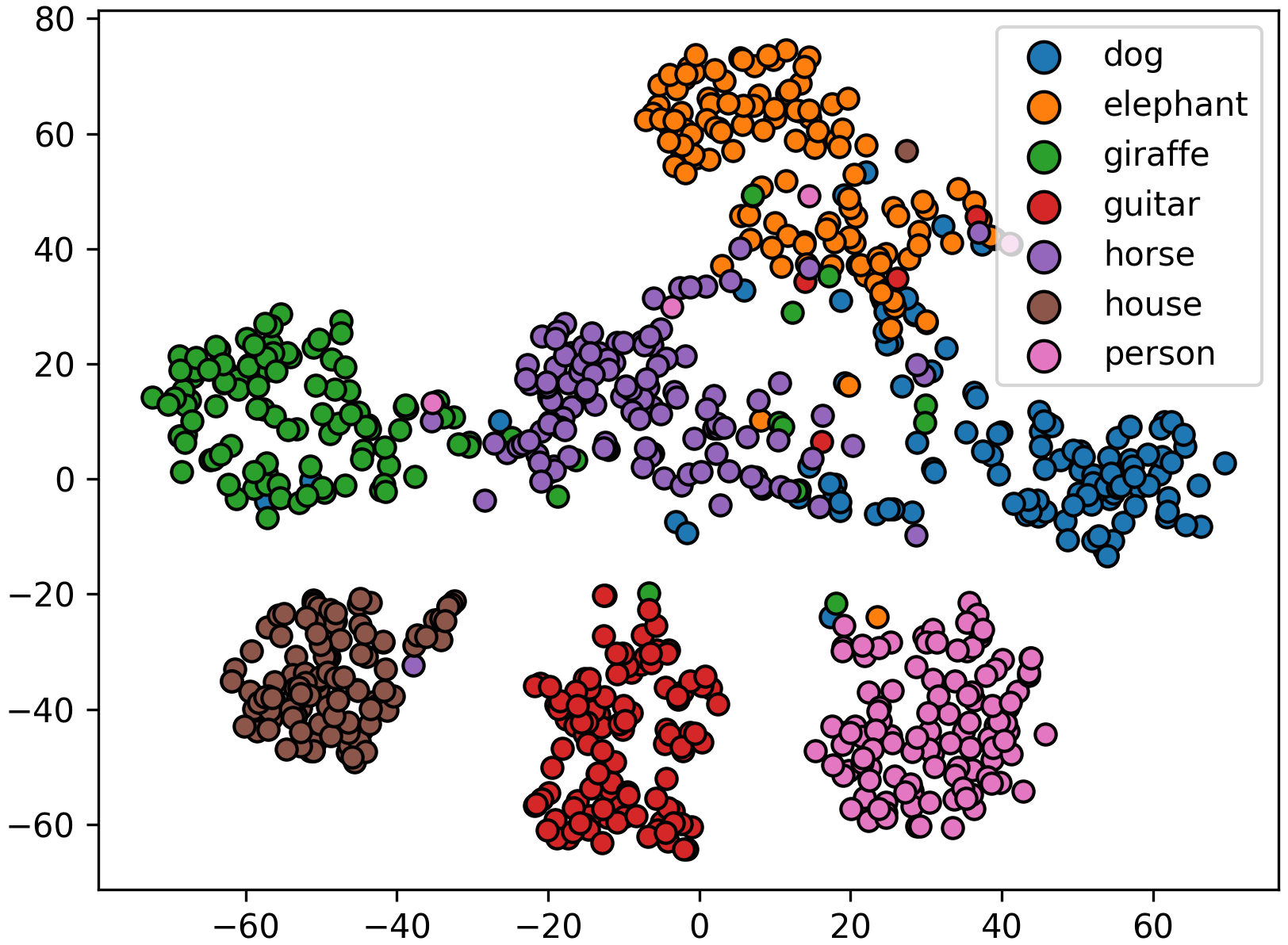}
\end{minipage}%
}%
\subfigure[domain-wise]{
\begin{minipage}[t]{0.5\columnwidth}
\centering
\includegraphics[width=1.55in]{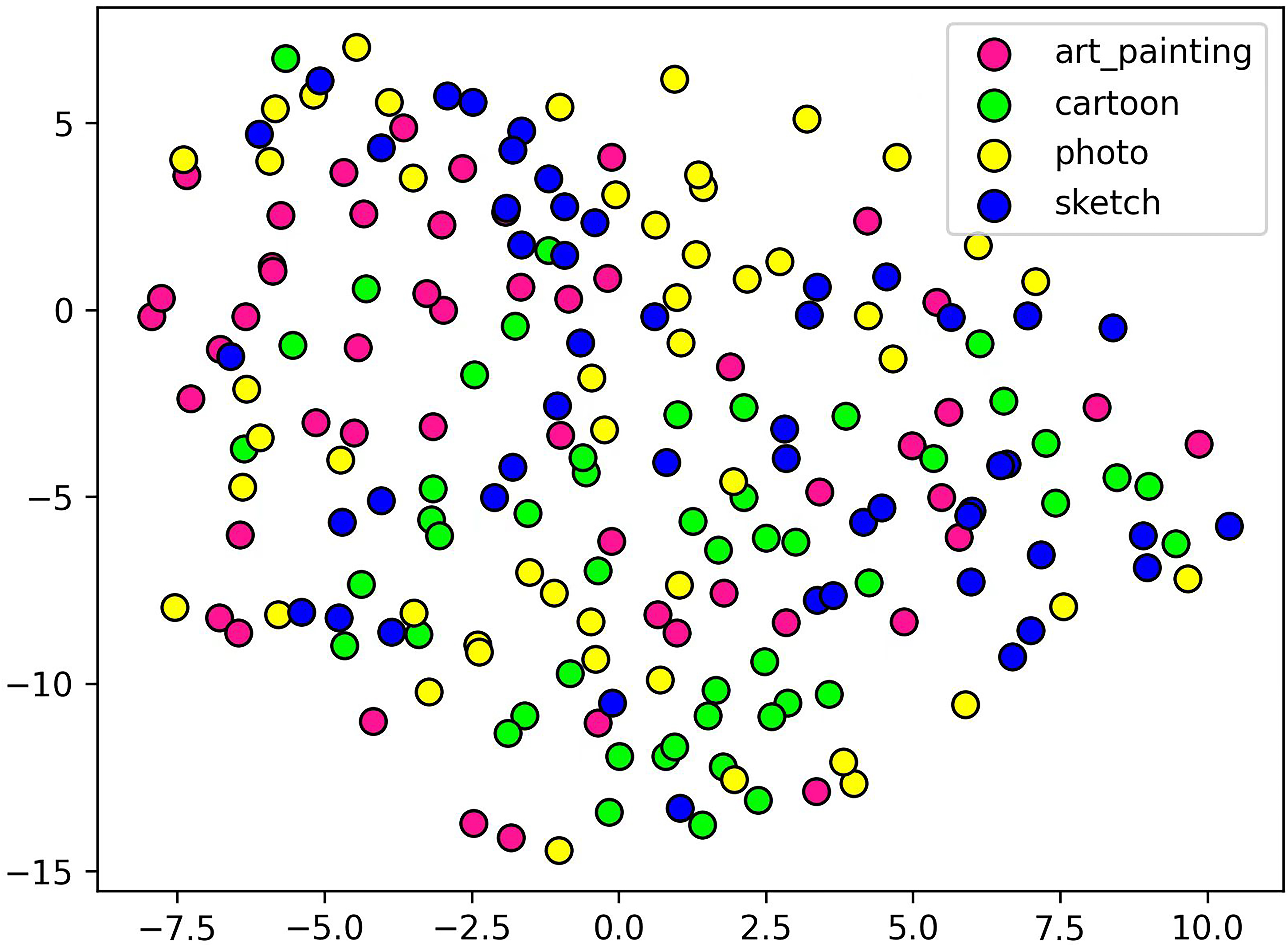}
\end{minipage}%
}%
\centering
\vskip -0.12in
\caption{\textbf{Feature distribution of the proposed PoER.} A vanilla ResNet-18 \cite{resnet} is trained on PACS \cite{PACS} dataset, and the above images show the category-wise and domain-wise feature distribution. We remove the conventional classification head and perform clustering using the outputted feature. (a): Feature distribution across different categories. It is clear that the final outputs of PoER are pure label-related representations. (b): Feature distribution across different domains. PoER makes the feature extractor aware of the characteristic of both label-related and domain-related information first and filters out the domain appearance progressively with the stacked convolutional blocks, achieving feature alignment for better generalization ability.}
\label{fig:top_figure}
\vskip -0.2in
\end{figure}

Commonly, a straightforward way to deal with the domain generalization problem is to collect as much data as possible from diverse domains for training. However, this solution is costly and impractical in some fields, and it is actually inevitable that the deep neural networks deployed in open-world scenarios will encounter out-of-distribution (OOD) inputs that are never exposed in the training phase no matter how much data you collect. Except for the aforementioned solution, the schemes aiming to mitigate the negative effects of domain shifts can be roughly divided into three categories, which are data augmentation schemes, representation learning techniques, and optimization methods. The data augmentation schemes such as \cite{DDAIG, g_domain} and \cite{mixstyle} mainly generate auxiliary synthetic data for training, improving the robustness and generalization ability of classifiers. The representation learning represented by feature alignment \cite{feature_alignment} enforces the neural networks to capture domain-invariant features, removing the irrelevant correlations between objects and background (\emph{i.e.,} domain appearance). There are adequate works of literature in this line of research, such as \cite{cross_grad, domain_confusion, invariant_feature} and \cite{stable_net}. Apart from the previously mentioned research lines, as stated in \cite{OODG_survey}, optimization methods that are both model agnostic and data structure agnostic are established to guarantee the worst-case performance under domain shifts. From the overall view, our proposed PoER belongs to the representation learning methods, but it deals with the DG problem from a brand new perspective. Instead of enforcing the neural network to generate domain-invariant features directly as presented in existing methods, PoER makes the neural networks capture both label-related and domain-related information explicitly first, and then distills the domain-related features out progressively, which in turn promotes the generation of domain-invariant features.

As the saying goes, \emph{know yourself and know your enemy, and you can fight a hundred battles with no danger of defeat}. The main drawback of existing representation learning methods is paying much attention to the generation of domain-invariant features before knowing the characteristics of the domain itself. By comparison, PoER makes the classifiers capture label-discriminative features containing domain information explicitly first in shallow layers, and with the distillation ability of the stacked convolutional blocks, filters the irrelevant correlations between objects and domains out. From the perspective of potential energy, PoER enforces the features with identical domain labels or category labels to have lower energy differences (\emph{i.e.,} pair-potential) in shallow layers and vice versa. Further, in deeper convolutional layers, PoER penalizes the classifiers if features with identical category labels are pushed far away from each other no matter domain labels, achieving domain decorrelation for better generalization ability. The key contributions of this paper are summarized as follows:

\begin{itemize}
\item A plug-and-play regularization term, namely PoER, is proposed to mitigate the negative effects of domain shifts. PoER is parameters-free and training-stable, which can be effortlessly combined with the mainstream neural network architectures, boosting the generalization ability of conventional classifiers.

\item PoER reports superior performance on domain generalization benchmarks, reducing the classification error by at least 1.20\% compared with existing methods. PoER is scalable to the size of datasets and images.

\item We tackle the domain generalization problem from a brand new perspective, \emph{i.e.,} potential energy ranking, wishing to introduce more insights to DG task.
\end{itemize}

\section{Related Work}
\subsubsection{Domain Augmentation Schemes.} This line of research argues that diverse training data is the key to more generalizable neural networks. In DDAIG \cite{DDAIG}, the adversarial training schemes are used for generating images from unseen domains that served as the auxiliary data, boosting the generalization ability of conventional neural networks. In MixStyle \cite{mixstyle}, the InstanceNorm \cite{IN} and AdaIN \cite{AdaIN} are used for extracting domain-related features. The mixing operation between these features results in representations from novel domains, increasing the domain diversity of training data (\emph{i.e.,} source domains). More recently, Style Neophile \cite{style_neophile} synthesizes novel styles constantly during training, addressing the limitation and maximizing the benefit of style augmentation. In SSAN \cite{style_shuffle}, different content and style features are reassembled for a stylized feature space, expanding the diversity of labeled data. Similarly, EFDM \cite{EFDM} proposes to match the empirical Cumulative Distribution Functions (eCDFs) of image features, mapping the representation from unseen domains to the specific feature space.

\subsubsection{Domain-invariant Representation Learning.} This is another research line for dealing with the domain generalization problem from the perspective of representation learning. In \cite{feature_alignment}, the deep features are promoted to be discriminative for the main learning task and invariant with respect to the shift across domains. In \cite{domain_confusion, cross_grad}, the neural networks are guided to extract domain-invariant features which can mitigate the effects of domain shifts. In \cite{depth_estimation}, self-supervised manners are extended for learning domain-invariant representation in depth estimation, yielding better generalization ability. \cite{domain_density} obtains a domain-invariant representation by enforcing the representation network to be invariant under all transformation functions among domains. Also in LIRR \cite{semi_DA}, the main idea is to simultaneously learn invariant representations and risks under the setting of Semi-DA. More recently, in CCT-Net \cite{cctnet}, the author employs confidence weighted pooling (CWP) to obtain coarse heatmaps which help generate category-invariant characteristics, enabling transferability from the source to the target domain.

\begin{figure*}[t]
\centering
\includegraphics[width=16cm]{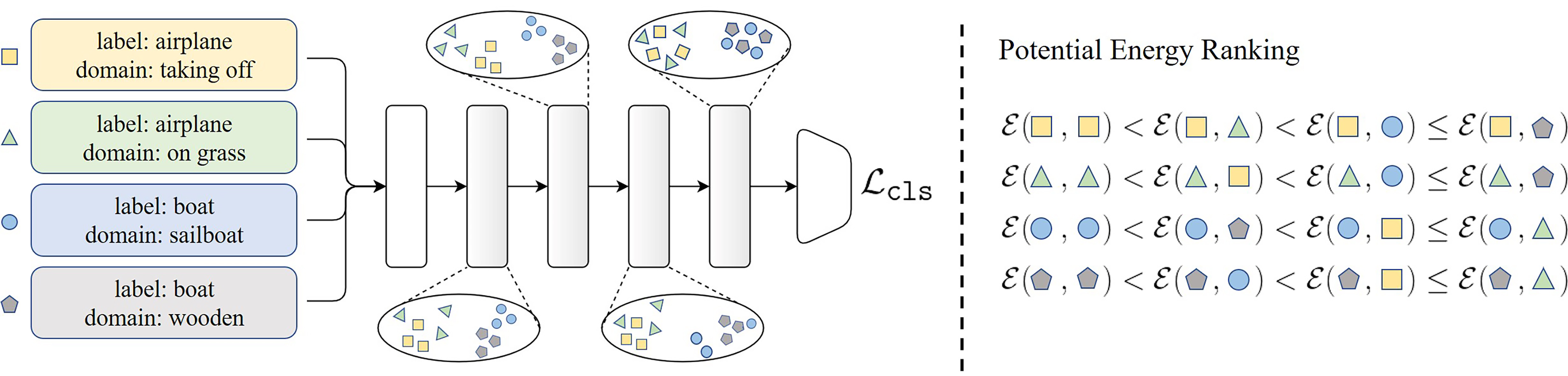}
\caption{\textbf{The proposed PoER framework.} In the shallow layers, neural networks extract feature representations containing both label-related and domain-related information, and PoER pushes the features with different domains far away from each other, making the neural networks aware the characteristic across domains. Following this, with the stacked convolutional blocks, PoER enforces the features within identical category labels close to each other progressively no matter domain labels, filtering out the irrelevant correlation between objects and domains. The distilled pure label-related feature is used for classification finally. In the image above, we use data within two categories and four domains as an example for depiction. $\mathcal{E}(,)$ indicates the pair-potential which describes the difference of potential energy between any given feature pairs.}
\label{nico}
\end{figure*}

\section{Preliminaries}
\subsubsection{Problem Statement.} Suppose $\mathcal{X}$, $\mathcal{D}$, and $\mathcal{Y}$ are the raw training data, domain labels, and category labels respectively. A classifier is defined as $f_\theta: \mathcal{X} \mapsto \mathcal{Y}$. Domain generalization aims to sample data only from the joint distribution $\mathcal{X}\times \mathcal{D} \times \mathcal{Y}$ for training while obtaining model $f_\theta$ which can generalize well to unseen domains. Feel free to use the domain labels or not during the training. Under the DG settings, $f_\theta$ has no access to the data from target domains, which is different from the DA task.

\subsubsection{Background of Potential Energy.} It is generally acknowledged that the energy is stored in objects due to their position in the field, namely potential energy. Pair potential is a function that describes the difference in potential energy between two given objects. A cluster of objects achieves stability when their pair potential between each other is fairly low. Inspired by this principle, we treat the representation and feature space of neural networks as objects and the potential field respectively. The classifier is expected to achieve stability where the energy difference (\emph{i.e.,} pair-potential) between the same domains and the same labels is lower. Based on this fact, we enforce the neural network to capture the representation explicitly containing the domain and the label information, and with the stacked convolutional blocks, filtering out the irrelevant correlation between label-related objects and the appearance. This means the pair potential is increasing across data with different category labels while decreasing across domains with identical category labels.

\section{Methods}
In this section, we detail our energy-based modeling first, elaborating PoER's methodology. Following that, we give the training and inference pipeline for reproducing in clarity.
\subsection{Energy-based Modeling}
We treat the feature space as a potential field $\mathcal{F}$, and the feature map from different layers is described as $f$. Formally, $d(,)$ is a metric function that measures the distance (\emph{i.e.,} potential difference) between any given feature pairs. Picking $L_2$ distance as the measurement, the potential difference is obtained as:
\begin{equation}
d(f^a,f^b)=\sqrt{\sum_{i=1}^{m}(f^a_i-f^b_i)^2}
\label{eq:potential-difference}
\end{equation}
\noindent
where $f^a$ and $f^b$ represent the flattened feature maps extracted from the identical layers of neural networks, and $m$ is their dimensionality. With an energy kernel $\mathcal{E}(,)$, the potential difference in Eq.(\ref{eq:potential-difference}) is mapped to the pair-potential as shown below:
\begin{equation}
\mathcal{E}(f^a,f^b)=\exp(\beta \cdot d(f^a, f^b))-1
\label{eq:pair-potential}
\end{equation}
where $\beta$ is a hyper-parameter, and it is equal to 1 by default. In the shallow layers, PoER pushes the features from identical categories and domains close to each other while keeping that from different categories or domains away, making the neural networks aware of the characteristics of domains and objects. To meet this awareness, PoER employs margin-based ranking loss to supervise the discrimination across different domains and categories. Intuitively, it is straightforward to accept that the features with identical category and domain labels should have lower pair-potential than those with different category or domain labels.

Suppose $f^{ij}$ is a feature representation from category $i$ and domain $j$, and $f^{pq}$ indicates the feature with category label $p$ and domain label $q$. Noting that $i\neq p$ and $j\neq q$. To formalize the idea of PoER, we build the ranking order across these features as follows:
\begin{equation}
\mathcal{E}(f^x,f^{ij}) < \mathcal{E}(f^x,f^{iq}) < \mathcal{E}(f^x,f^{pj}) < \mathcal{E}(f^x,f^{pq})
\label{eq:ranking-order}
\end{equation}
where $f^x$ indicates another feature with the identical category and domain label as $f^{ij}$. With the margin-based ranking loss, we combine the feature pairs above to calculate the pair-wise ranking loss. Formally, we have:
\begin{align}
\mathcal{L}_1&=\max(0, \mathcal{E}(f^x,f^{ij}) - \mathcal{E}(f^x,f^{iq}) + \delta)\\
\mathcal{L}_2&=\max(0, \mathcal{E}(f^x,f^{iq}) - \mathcal{E}(f^x,f^{pj}) + \delta) \\ 
\mathcal{L}_3&=\max(0, \mathcal{E}(f^x,f^{pj}) - \mathcal{E}(f^x,f^{pq}) + \delta)
 \label{eq:ranking_loss}
\end{align}
where $\delta$ indicates a non-negative margin, and we set $\delta$ to 0 in our experiments. The complete ranking loss in shallow layers is depicted as $\mathcal{L}_\texttt{rank}=\mathcal{L}_1+\mathcal{L}_2+\mathcal{L}_3$.

With the stacked convolutional blocks, PoER filters the domain-related information out progressively. In deeper layers, we enforce the features distribute closely intra-class while discretely inter-class, ignoring the domain labels. We use $f^{i*}$ and $f^{p*}$ to depict the data with category label $i$ and $p$ respectively. Therefore, given feature $f^x$ from category $i$, the cluster loss is formulated as follows:
\begin{equation}
\mathcal{L}_\texttt{cluster}=\exp(\mathcal{E}(f^x,f^{i*}) - \mathcal{E}(f^x,f^{p*}))
\label{eq:cluster-loss}
\end{equation}

The PoER regularization is the sum of aforementioned loss functions, saying $\mathcal{L}_\texttt{PoER}=\mathcal{L}_\texttt{rank}+\mathcal{L}_\texttt{cluster}$. Moreover, the aforementioned regularization term can be easily calculated within each batch of the training data since all of the combinations stated above exist.

\subsection{Distance-based Classification}
We classify the data in GCPL \cite{gcpl} manner since we regularize the features in neural networks. $f$ is the feature from the penultimate layer of the discriminative model such as conventional ResNet-18 with shape $m$, and $M$ is the learnable prototypes with shape $k\times n\times m$, where $k$ and $n$ are the numbers of classes and prototypes each class. The $L_2$ distance is calculated between $f$ and $M$ along the last dimension of prototypes. The distance matrix is obtained as $d(f,M)$ in shape $k\times n$. For calculating the category cross-entropy loss, we pick the minimal distance within prototypes of each class, saying $d$ is acquired by selecting the minimum value along the last dimension of $d(f,M)$ in shape $k$. Formally, the predicted probability that given data belongs to category $i$ is built as:
\begin{equation}
p_i=\frac{\exp(-d_i)}{\sum_{j=1}^{k}\exp(-d_j)}
\label{eq:pi}
\end{equation}

Suppose $\mathbb{I}(y=i)$ is an indicator function that equals 1 if and only if the corresponding label of $f$ is $i$ otherwise 0, and $y$ is the category label of feature $f$. Therefore, the distance-based classification loss is then formulated as:
\begin{equation}
\mathcal{L}_\texttt{cls}=-\sum_{i=1}^{k} \mathbb{I}(y=i) \cdot \log(p_i)
\label{eq:cls-loss}
\end{equation}

The overall loss function of distance-based classification with PoER is in the form as:
\begin{equation}
\mathcal{L}=\mathcal{L}_\texttt{cls}+\alpha \mathcal{L}_\texttt{PoER}
\label{eq:loss}
\end{equation}
where $\alpha$ is a hyper-parameter to balance the regularization term and classification error.

\subsection{Training and Inference}
We detail the training and inference pipeline in this section for easy reproduction and a clear understanding. As stated in the previous section, $\mathcal{X}$, $\mathcal{D}$, and $\mathcal{Y}$ are the space of image data, domain labels, and category labels. We sample data $(x,d,y)$ from the joint distribution of $\mathcal{X} \times \mathcal{D} \times \mathcal{Y}$ for training. Suppose $h(\cdot)$ is a feature extractor that returns the flattened features of each block in neural networks. The feature extractor has no more need of a classification head since we employ the distance-based cluster manner for identification. The training and inference processes are summarized in Algorithm \ref{algo:poer}. It is worth noting that PoER performs regularization on features from all blocks while the classification is achieved only with the features from the last block of $h(\cdot)$. 

\setlength{\textfloatsep}{5pt}
\begin{algorithm}[t]
    \caption{Potential energy ranking for DG task.}
    \label{algo:poer}
    \LinesNumbered
    \KwIn{training data $\mathcal{X} \times \mathcal{D} \times \mathcal{Y}$, neural network $h(\cdot)$}
    \KwOut{trained neural network $h(\cdot)$}
    \While {\texttt{Training}}{
			{Sample a batch data $\{x,d,y\}$ from $\mathcal{X} \times \mathcal{D} \times \mathcal{Y}$\;
			Get the features from each block: $f_s=h(x)$\;
			Calculate $\mathcal{L}_\texttt{rank}$ for features from the first three blocks in $f_s$ with pair-wise manner\;
            Calculate $\mathcal{L}_\texttt{cluster}$ for features from the left blocks in $f_s$, including the last one\;
			Calculate the $\mathcal{L}_{\texttt{cls}}$ with feature from the last block in $f_s$, summing them up as Eq.(\ref{eq:loss})\;
			Update the parameters of $h(\cdot)$ with gradient descent method.}
}

    \While {\texttt{Inference}}{
            Sample $x$ from the testing set\;
            Get feature $f$ from the last block of $h(\cdot)$\;
			Calculate distance between $f$ and prototypes $M$ with Eq.(\ref{eq:potential-difference})\;
			Classify the given data using Eq.(\ref{eq:pi}).
}
\end{algorithm}

\section{Experiments}
\subsection{Dataset}
We consider four benchmarks to evaluate the performance of our proposed PoER, namely PACS \cite{PACS}, VLCS \cite{vlcs}, Digits-DG \cite{mixstyle}, and Office-Home \cite{office-home}. On the NICO \cite{nico} dataset, we only report the limited results of some new methods we collected. The datasets mentioned below can be downloaded at Dassl \cite{dassl}, which is a testing bed including many DG methods.

\noindent
\textbf{PACS} contains images with shape $227\times 227 \times 3$ in RGB channel, belonging to 7 categories within 4 domains which are \textbf{P}hoto, \textbf{A}rt, \textbf{C}artoon, and \textbf{S}ketch. Under DG settings, the model has no access to the target domain, and therefore the dataset is split into three parts used for training, validation, and test. We use the split file provided in EntropyReg \cite{EntropyReg}. The training and validation sets are data from the source domains while the test set is sampled from the target domain. We pick classifiers based on the validation metric for reporting the test results.

\noindent
\textbf{Office-Home} contains images belonging to 65 categories within 4 domains which are artistic, clip art, product, and the real world. Following DDAIG \cite{DDAIG}, we randomly split the source domains into 90\% for training and 10\% for validation, reporting the metrics on the leave-one-out domain using the best-validated model.

\noindent
\textbf{Digits-DG} is a mixture of 4 datasets, namely MNIST \cite{MNIST}, MNIST-M \cite{feature_alignment}, SVHN \cite{svhn}, and SYN \cite{feature_alignment}. All images are resized into $32 \times 32 \times 3$. The reported metrics use the leave-one-domain-out manner for evaluation.

\noindent
\textbf{VLCS} contains images from 5 categories within 4 domains which are Pascal \textbf{V}OC2007 \cite{voc}, \textbf{L}abelMe \cite{labelme}, \textbf{C}altech \cite{caltech}, and \textbf{S}UN09 \cite{sun}. We randomly split the source domains into 70\% for training and 30\% for validation following \cite{vlcs}, reporting metrics on the target domain using the best-validated classifier.

\noindent
\textbf{NICO} consists of natural images within 10 domains, 8 out of which are treated as the source and 2 as the target. Following \cite{stable_net}, we randomly split the data into 90\% for training and 10\% for validation, reporting metrics on the left domains with the best-validated model.

\subsection{Evaluation Protocol}
We report top-1 classification accuracy on the aforementioned datasets. For avoiding occasionality, each setting is measured with 5 runs. We also give the 95\% confidence intervals calculated with $\mu \pm 1.96\frac{\sigma}{\sqrt{k}}$, where $\mu$, $\sigma$, and $k$ are the mean, standard deviation, and runs of the top-1 accuracy. A part of previous methods report no 95\% confidence intervals, and therefore, we give the top-1 classification accuracy.

\subsection{Experimental Setup}
We use ResNet-18 \cite{resnet} without the classification head as the feature extractor $h(\cdot)$, and the backbone is pre-trained on ImageNet \cite{imagenet}. $h(\cdot)$ has 5 blocks including the top convolutional layer. We reduce the dimension of the outputted feature of ResNet from 512 to 128 with a linear block. For summarize, our $h(\cdot)$ returns 6 flattened features in total. The learning rate starts from 1e-4 and halves every 70 epochs. The batch size is set to 128. The hyper-parameter $\alpha$ in Eq.(\ref{eq:loss}) is set to 0.1 during the first 70 epochs otherwise 0.2. Only the \texttt{RandomHorizontalFlip} and \texttt{ColorJitter} are adopted as the data augmentation schemes. The AdamW optimizer is used for training. The mean-std normalization is used based on the ImageNet statistics. GCPL \cite{gcpl} uses the same settings as stated above, and all other methods employ the default official settings. We store the models after the first 10 epochs based on the top-1 accuracy on the validation set. The number of prototypes $n$ is set to 3.

\subsection{Comparisons with State-of-the-Arts}
We report the main results of domain generalization on common benchmarks in this section. If not specified, the ResNet-18 \cite{resnet} is adopted as the backbone across different techniques. We use \textbf{Avg.} to represent the average top-1 classification accuracy over different domains.
\subsubsection{Leave-one-domain-out results on PACS.} Since we only collect limited results with 95\% confidence intervals, we report the mean top-1 accuracy over 5 runs. Methods shown in the upper part of Table \ref{tab:pacs} use AlexNet \cite{alexnet} as the backbone while the following part in gray background uses ResNet-18 \cite{resnet}. The vanilla counterpart of PoER is GCPL \cite{gcpl}. PoER improves the top-1 classification accuracy up to 2.32\% and 0.48\% compared to its vanilla counterpart and the existing \emph{state-of-the-art} methods. Methods shown below are arranged in the decreasing order of top-1 accuracy. A, C, P, and S in Table \ref{tab:pacs} indicate Art, Cartoon, Photo, and Sketch.

\setlength{\tabcolsep}{2pt}
\begin{table}[t]\footnotesize
\begin{center}
\begin{tabular}{l|cccc|c}
\hline
Methods  & A. & C. & P. & S. & Avg. \\ \hline
D-MATE \cite{D-MATE}& 60.27 & 58.65 & 91.12 & 47.68 & 64.48\\
M-ADA \cite{M-ADA} & 61.53 & 68.76 & 83.21 & 58.49 & 68.00\\
DBADG \cite{PACS} & 62.86 & 66.97 & 89.50 & 57.51 & 69.21\\
MLDG \cite{MLDG} & 66.23 & 66.88 & 88.00 & 58.96 & 70.01\\
Feature-critic \cite{feature-critic} & 64.89 & 71.72 & 89.94 & 61.85 & 71.20\\
CIDDG \cite{CIDDG} & 66.99 & 68.62 & 90.19 & 62.88 & 72.20\\
MMLD \cite{mmld} & 69.27 & 72.83 & 88.98 & 66.44 & 74.38\\
MASF \cite{masf} & 70.35 & 72.46 & 90.68 & 67.33 & 75.21 \\
EntropyReg \cite{EntropyReg} & 71.34 & 70.29 & 89.92 & 71.15 & 75.67\\\rowcolor{gray!15}
MMD-AAE \cite{MMD-AAE} & 75.20 & 72.70 & 96.00 & 64.20 & 77.03 \\\rowcolor{gray!15}
CCSA \cite{CCSA} & 80.50 & 76.90 & 93.60 & 66.80 & 79.45\\\rowcolor{gray!15}
ResNet-18 \cite{resnet} & 77.00 & 75.90 & 96.00 & 69.20 & 79.53\\\rowcolor{gray!15}
StableNet \cite{stable_net} & 80.16 & 74.15 & 94.24 & 70.10 & 79.66\\\rowcolor{gray!15}
JiGen \cite{JiGen} & 79.40 & 75.30 & 96.00 & 71.60 & 80.50\\\rowcolor{gray!15}
CrossGrad \cite{cross_grad} & 79.80 & 76.80 & 96.00 & 70.20 & 80.70\\\rowcolor{gray!15}
DANN \cite{feature_alignment} & 80.20 & 77.60 & 95.40 & 70.00 & 80.80\\\rowcolor{gray!15}
Epi-FCR \cite{epi-fcr} & 82.10 & 77.00 & 93.90 & 73.00 & 81.50\\\rowcolor{gray!15}
MetaReg \cite{meta-reg} & 83.70 & 77.20 & 95.50 & 70.30 & 81.70\\\rowcolor{gray!15}
GCPL \cite{gcpl} & 82.64 & 75.02 & 96.40 & 73.36 & 81.86\\\rowcolor{gray!15}
EISNet \cite{EISNet} & 81.89 & 76.44 & 95.93 & 74.33 & 82.15\\\rowcolor{gray!15}
L2A-OT \cite{L2A-OT} & 83.30 & 78.20 & 96.20 & 73.60 & 82.83\\\rowcolor{gray!15}
MixStyle \cite{mixstyle} & 84.10 & \textbf{78.80} & 96.10 & 75.90 & 83.70\\\rowcolor{gray!15} \hline
PoER (Ours) & \textbf{85.30} & 77.69 & \textbf{96.42} & \textbf{77.30} & \textbf{84.18}\\ \hline
\end{tabular}
\caption{Leave-one-domain-out results on PACS dataset without 95\% confidence intervals. The methods in gray background use ResNet-18 as the backbone while other methods employ AlexNet for feature extraction.}
\label{tab:pacs}
\end{center}
\end{table}

\subsubsection{Leave-one-domain-out results on OfficeHome dataset.} We report the mean top-1 accuracy and 95\% confidence interval results on OfficeHome. Some of the following results are from DDAIG \cite{DDAIG}. We use the same method as stated in DDAIG to split the source domains into 90\% for training and 10\% for validation. The images in OfficeHome are colorful in the RGB channel whose scale scatters from $18\times 18$ pixels to $6500\times 4900$ pixels. The short edge of all images is resized to 227 first, maintaining the aspect ratio, and then the training inputs are obtained through \texttt{RandomResizedCrop} with shape 224. In Table \ref{tab:office-home}, CCSA, MMD-AAE, and D-SAM are from \cite{CCSA}, \cite{MMD-AAE}, and \cite{D-SAM}, and other methods have been introduced before. As stated in the previous section, the vanilla counterpart of PoER is GCPL. It can be found that PoER reduces the classification error by a clear margin of 1.3\% and 1.2\% compared to its vanilla counterpart and the \emph{state-of-the-art} method DDAIG.

\setlength{\tabcolsep}{1.9pt}
\begin{table}[t]
\begin{center}
\begin{tabular}{l|cccc|c}
\hline
Method & Artistic & Clipart & Product & Real World & Avg. \\ \hline
ResNet-18 & 58.9$\pm$.3 & 49.4$\pm$.1 & 74.3$\pm$.1 & 76.2$\pm$.2 & 64.7    \\
CCSA & 59.9$\pm$.3 & 49.9$\pm$.4 & 74.1$\pm$.2 & 75.7$\pm$.2 & 64.9    \\
MMD-AAE & 56.5$\pm$.4 & 47.3$\pm$.3 & 72.1$\pm$.3 & 74.8$\pm$.2 & 62.7    \\
CrossGrad & 58.4$\pm$.7 & 49.4$\pm$.4 & 73.9$\pm$.2 & 75.8$\pm$.1 & 64.4    \\
D-SAM & 58.0  & 44.4  & 69.2  & 71.5  & 60.8    \\
JiGen  & 53.0 & 47.5 & 71.5 & 72.8 & 61.2    \\
GCPL & 58.3$\pm$.1 & 51.9$\pm$.1 & 74.1$\pm$.2 & 76.7$\pm$.1 & 65.3    \\
DDAIG & \textbf{59.2$\pm$.1} & 52.3$\pm$.3 & 74.6$\pm$.3 & 76.0$\pm$.1 & 65.5    \\ \hline 
PoER (ours) & 59.1$\pm$.2 & \textbf{53.4$\pm$.3} & \textbf{74.9$\pm$.2} & \textbf{79.1$\pm$.3} & \textbf{66.6}    \\ \hline
\end{tabular}
\caption{Leave-one-domain-out results on OfficeHome dataset with 95\% confidence intervals. No confidence intervals are reported in the original paper of D-SAM and JiGen.}
\label{tab:office-home}
\end{center}
\end{table}

\subsubsection{Domain generalization results on NICO dataset.} NICO is different from the aforementioned dataset. It consists of two super-categories, namely Animal and Vehicle, including 19 sub-classes in total. Moreover, the domains of each sub-class are different from each other. In a nutshell, NICO contains 19 classes belonging to 65 domains. For each class, we randomly select 2 domains as the target while the left 8 domains are treated as the source. Within source domains, we further split the data into 90\% for training and 10\% for validation. The metrics are reported with the best-validated models on target domains. RSC indicates the algorithm from \cite{RSC}. No pre-trained weights are used in Table \ref{tab:nico}. PoER reports the superior performance by a remarkable margin of 2.83\% compared to the existing methods.

\setlength{\tabcolsep}{5.1pt}
\begin{table}[htpb]
\begin{center}
\begin{tabular}{lcccc}
\hline
\multicolumn{1}{c}{}     & M-ADA & MMLD & ResNet-18 & \multirow{4}{*}{\begin{tabular}[c]{@{}c@{}}PoER (ours)\\ \textbf{62.62}\end{tabular}} \\ \cline{2-4}
\multicolumn{1}{c}{NICO} & 40.78 & 47.18   & 51.71     &                                                                              \\ \cline{1-4}
                         & JiGen & RSC     & StableNet &                                                                              \\ \cline{2-4}
NICO                     & 54.42 & 57.59   & 59.76     &                                                                              \\ \hline
\end{tabular}
\caption{Domain generalization results on NICO dataset.}
\label{tab:nico}
\end{center}
\end{table}

\subsubsection{Leave-one-domain-out results on Digits-DG dataset.} Digits-DG consists of four different datasets containing digits with different appearances. Following DDAIG \cite{DDAIG}, all images are resized to $32\times 32$ with RGB channel. For the MNIST dataset, we replicate the gray channel three times to construct the color images. As stated in \cite{DDAIG}, we randomly pick 600 images for each class in these four datasets. Images are split into 90\% for training and 10\% for validation. The leave-one-domain-out protocol is used for evaluated the domain generalization performance. All images in the left domain are tested for reporting metrics. Table \ref{tab:digits-dg} reveals the 95\% confidence intervals of PoER and its comparisons with some existing domain generalization methods. It is clear to see that PoER surpasses previous techniques on most domains by a large margin, reducing the classification error up to 4.23\% and achieving newly \emph{state-of-the-art} domain generalization performance with only a vanilla ResNet-18 backbone. All methods shown in Table \ref{tab:digits-dg} are presented in previous sections. 

\setlength{\tabcolsep}{2.2pt}
\begin{table}[t]
\begin{center}
\begin{tabular}{l|cccc|c}
\hline
Method      & MNIST       & MNIST-M     & SVHN        & SYN         & Avg. \\ \hline
ResNet-18   & 95.8$\pm$.3 & 58.8$\pm$.5 & 61.7$\pm$.5 & 78.6$\pm$.6 & 73.7 \\
CCSA        & 95.2$\pm$.2 & 58.2$\pm$.6 & 65.5$\pm$.2 & 79.1$\pm$.8 & 74.5 \\
MMD-AAE     & 96.5$\pm$.1 & 58.4$\pm$.1 & 65.0$\pm$.1 & 78.4$\pm$.2 & 74.6 \\
CrossGrad   & 96.7$\pm$.1 & 61.1$\pm$.5 & 65.3$\pm$.5 & 80.2$\pm$.2 & 75.8 \\
GCPL       & 96.3$\pm$.1 & 58.7$\pm$.5 & 70.2$\pm$.3 & 80.5$\pm$.3 & 76.4 \\
DDAIG       & 96.6$\pm$.2 & \textbf{64.1$\pm$.4} & 68.6$\pm$.6 & 81.0$\pm$.5 & 77.6 \\
\hline
PoER (ours) & \textbf{97.2$\pm$.4} & 60.1$\pm$.3 & \textbf{75.6$\pm$.4} & \textbf{94.4$\pm$.3} & \textbf{81.8} \\ \hline
\end{tabular}
\caption{Leave-one-domain-out results on Digits-DG with 95\% confidence intervals.}
\label{tab:digits-dg}
\end{center}
\end{table}

\subsubsection{Leave-one-domain-out results on VLCS dataset.} VLCS consists of four common datasets. All methods shown in Table \ref{tab:vlcs} can be found in Table \ref{tab:pacs} in detail. VOC indicates the Pascal VOC dataset. A part of the following results is from StableNet \cite{stable_net}. Following \cite{EntropyReg}, the leave-one-domain-out protocol is used for evaluation. The source domains are split into 70\% for training and 30\% for validation. The best-validated model reports domain generalization performance on all images from the left target domain. Since the size of images in VLCS is varied from each other, we thus resize the short edge of images to 227 while keeping their aspect ratio and then randomly crop squares with shape 224 for training. Results depicted in Table \ref{tab:vlcs} reveal that PoER gives a better classification accuracy surpassing other methods with a large improvement, saying 1.44\% outperforming the current techniques on Caltech.

\setlength{\tabcolsep}{4pt}
\begin{table}[t]
\begin{center}
\begin{tabular}{l|cccc|c}
\hline
Method      & VOC & LabelMe & Caltech & SUN09 & Avg.  \\ \hline
DBADG       & 69.99      & 63.49   & 93.64   & 61.32 & 72.11 \\
ResNet-18   & 67.48      & 61.81   & 91.86   & 68.77 & 72.48 \\
JiGen       & 70.62      & 60.90   & 96.93   & 64.30 & 73.19 \\
MMLD        & 71.96      & 58.77   & 96.66   & 68.13 & 73.88 \\
CIDDG       & 73.00      & 58.30   & 97.02   & 68.89 & 74.30 \\
EntropyReg  & 73.24      & 58.26   & 96.92   & 69.10 & 74.38 \\
GCPL        & 67.01      & 64.84   & 96.23   & 69.43 & 74.38 \\
RSC         & \textbf{73.81}      & 62.51   & 96.21   & 72.10 & 76.16 \\
StableNet   & 73.59      & 65.36   & 96.67   & \textbf{74.97} & \textbf{77.65} \\ \hline
PoER (ours) & 69.96      & \textbf{66.41}  & \textbf{98.11}   & 72.04 & 76.63      \\ \hline
\end{tabular}
\caption{Leave-one-domain-out results on VLCS dataset. PoER gets lower metrics on Pascal VOC and SUN09 while reporting superior performance on Caltech.}
\label{tab:vlcs}
\end{center}
\end{table}

\subsection{Ablation Study}
\subsubsection{Ablation on the weight of PoER $\alpha$.} In Eq.(\ref{eq:loss}), we introduce a hyper-parameter $\alpha$ for balancing the classification loss and energy ranking loss. We set this parameter from 0.0 to 0.9 with a step 0.1 for testing its sensitivity. To clarify, $\alpha$ equals to 0 indicates the vanilla counterpart of PoER, \emph{i.e.,} GCPL. We use the PACS benchmark and set the number of prototypes $k$ to 3. The ablation results are shown in Table \ref{tab:ablation-a}. From Table \ref{tab:ablation-a}, we find that the performance is better when setting $\alpha$ to 0.2, considering the training stability, we set $\alpha$ to 0.1 in the first 70 epochs and otherwise 0.2.

\setlength{\tabcolsep}{6.5pt}
\begin{table}[t]
\begin{center}
\begin{tabular}{lccccc}
\hline
$\alpha$ & 0.0   & 0.1   & 0.2   & 0.3   & 0.4   \\ \cline{2-6} 
PACS Avg. & 81.86 & 83.92 & \textbf{84.20} & 84.16 & 84.16 \\ \hline
$\alpha$ & 0.5   & 0.6   & 0.7   & 0.8   & 0.9   \\ \cline{2-6} 
PACS Avg. & 84.13 & 84.06 & 84.12 & 83.97 & 83.60 \\ \hline
\end{tabular}
\caption{Ablation results on hyper-parameter $\alpha$.}
\label{tab:ablation-a}
\end{center}
\end{table}

\subsubsection{Ablation on the number of prototypes $k$.} In Eq.(\ref{eq:pi}), we set the learnable prototypes as a tensor with shape $k\times n\times m$ where the $k$, $n$, and $m$ are the number of classes, the number of prototypes, and the dimensionality of the outputted feature from the last block. We test the impacts of the number of prototypes with respect to the classification performance from 1 to 10 with step 1 on PACS benchmark. From Table \ref{tab:ablation-k}, we find that more prototypes guarantee a better classification result to some extent. Considering both the performance and calculation efficiency, we set the number of prototypes $k$ to 3 default, leading to metrics which are marginally lower than the best one.

\setlength{\tabcolsep}{6.5pt}
\begin{table}[t]
\begin{center}
\begin{tabular}{lccccc}
\hline
$k$       & 1     & 2     & 3     & 4     & 5     \\ \cline{2-6} 
PACS Avg. & 82.40 & 83.97 & 84.18 & 84.12 & 84.07 \\ \hline
$k$       & 6     & 7     & 8     & 9     & 10    \\ \cline{2-6} 
PACS Avg. & 84.20 & 84.18 & 84.19 & 84.19 & \textbf{84.21} \\ \hline
\end{tabular}
\caption{Ablation results on the number of prototypes $k$.}
\label{tab:ablation-k}
\end{center}
\end{table}

\subsubsection{Ablation on the proposed loss functions.} In Eq.(\ref{eq:ranking_loss}) and Eq.(\ref{eq:cluster-loss}), we propose calculating ranking loss in the shallow layers while performing clustering regularization in the deeper layers. Taking ResNet-18 as an example, we extract features from the first three blocks (including the first convolutional layer) for calculating the ranking loss as shown in Eq.(\ref{eq:ranking_loss}) while the features extracted from the following blocks are treated as the deeper layers for getting clustering loss as shown in Eq.(\ref{eq:cluster-loss}). We test the combinations of different loss functions, namely $\mathcal{L}_\texttt{cls}$, $\mathcal{L}_\texttt{cls} + \alpha \mathcal{L}_\texttt{rank}$, $\mathcal{L}_\texttt{cls} + \alpha \mathcal{L}_\texttt{cluster}$, and $\mathcal{L}_\texttt{cls} + \alpha \mathcal{L}_\texttt{PoER}$. Noting that $\mathcal{L}_\texttt{PoER}$ actually equals to $\mathcal{L}_\texttt{rank}+\mathcal{L}_\texttt{cluster}$. Recall that $\mathcal{L}_\texttt{cls}$ indicates the vanilla counterpart of PoER, \emph{i.e.,} GCPL. We use the default settings of $\alpha$ and $k$ as stated in previous sections. The ablation results on different combinations of loss functions are presented in Table \ref{tab:ablation-loss}. Metrics are evaluated on the PACS benchmark. From Table \ref{tab:ablation-loss}, we find that both $\mathcal{L}_\texttt{rank}$ and $\mathcal{L}_\texttt{cluster}$ help to improve the domain generalization ability of conventional neural networks. Noting that $\mathcal{L}_\texttt{cluster}$ can be treated as the self-supervised manner for boosting domain generalization performance.

\setlength{\tabcolsep}{7.2pt}
\begin{table}[htpb]
\begin{center}
\begin{tabular}{lcc}
\hline
\emph{Loss}      & $\mathcal{L}_\texttt{cls}$ & $\mathcal{L}_\texttt{cls} + \alpha \mathcal{L}_\texttt{rank}$ \\ \cline{2-3} 
\textbf{PACS Avg.} & 81.86        & 83.72     \\ \hline
\emph{Loss}      & $\mathcal{L}_\texttt{cls} + \alpha \mathcal{L}_\texttt{cluster}$ & $\mathcal{L}_\texttt{cls} + \alpha \mathcal{L}_\texttt{PoER}$ \\ \cline{2-3} 
\textbf{PACS Avg.} & 82.60        & \textbf{84.18}     \\ \hline
\end{tabular}
\caption{Ablation results on different combinations of the proposed loss functions. Especially, as shown above, the group with $\mathcal{L}_\texttt{rank}$ surpasses $\mathcal{L}_\texttt{cluster}$ up to 1.12\%, suggesting the importance of domain ranking proposed in PoER.}
\label{tab:ablation-loss}
\end{center}
\end{table}

\noindent
\textbf{Ablation on the scalability of PoER.} We investigate the scalability of PoER among different model architectures. To verify this, we employ both the convolution- and attention-based models to perform experiments on NICO \cite{nico} dataset. Since the attention-based models are hard to converge, we resort to the pre-trained weights on ImageNet \cite{imagenet}. The convolution-based models are trained from scratch. Noting that we focus more on the improvement that PoER introduces to the corresponding vanilla counterparts. ResNet \cite{resnet} and DenseNet \cite{densenet} are the representatives of convolutionl models while Swin-Transformer (Tiny) \cite{swin} is that of attention-based models. Results in Table \ref{tab:ablation-model} demonstrate that PoER improves the top-1 classification accuracy over 3.55\% consistently, evidencing its scalability.

\setlength{\tabcolsep}{7.5pt}
\begin{table}[t]
\begin{center}
\begin{tabular}{lcccc}
\hline
Model       & RN-18     & RN-50     & DN-121     & Swin-T \\ \cline{2-5}
Avg. (\ding{55}) & 51.71 & 55.43 & 65.19 & 73.43\\ \rowcolor{gray!15}
Avg. (\ding{51}) & 62.62 & 64.10 & 69.75 & 76.98\\ \hline
\end{tabular}
\vskip -0.05in
\caption{Ablation results of different model architectures on NICO dataset. RN, DN, and Swin-T indicate ResNet, DenseNet, and Swin-Transformer. \ding{55} is the vanilla classifier while \ding{51} is the corresponding PoER version.}
\label{tab:ablation-model}
\end{center}
\end{table}

\subsection{Feature Visualization Results}
We provide the feature distribution within the identical category from the shallow layers to the deeper. 
Recall that in the shallow layers, PoER employs the ranking loss to make the model aware of the difference across different domains, thus we expect the distinction among varied domains within the identical category to become greater progressively.

\begin{figure}[htpb]
\centering
\includegraphics[width=8.5cm]{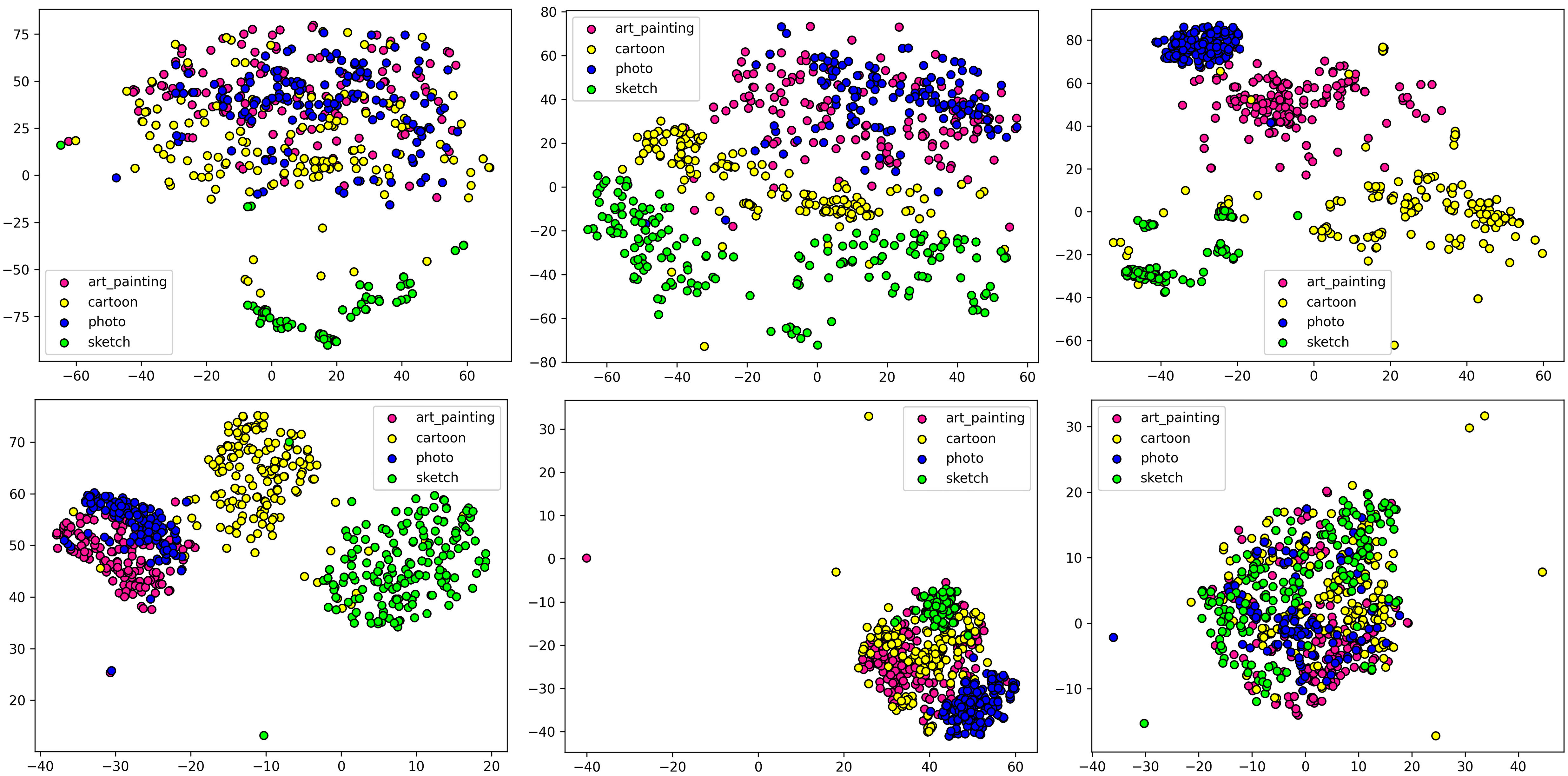}
\caption{\textbf{Visualization results of the feature distribution in each layer.} Images above are feature distributions with identical category label from block 1 to block 6 (from left to right and from top to bottom) successively on PACS.}
\label{vis}
\end{figure}

From the feature distribution of each block shown in Figure \ref{vis}, it can be concluded that in the first three blocks (the first row), PoER employs domain ranking loss to make the neural network aware of the differences across domains, separating the features from different domains. In the following blocks (the second row), PoER aims to filter the domain-related information out for clustering, making the features with identical category labels close together no matter the domains. As stated in the beginning, PoER learns the characters of each domain and category before generating domain-invariant features, laying the foundation for distilling pure label-related features.

\section{Conclusion}
This paper proposes using PoER to make the classifier aware of the characters of different domains and categories before generating domain-invariant features. We find and verify that PoER is vital and helpful for improving the generalization ability of models across domains. PoER reports superior results on sufficient domain generalization benchmarks compared to existing techniques, achieving \emph{state-of-the-art} performance. Insights of the proposed idea are given both statistically and visually. We hope the mentioned energy perspective can inspire the following works.

\section{Acknowledgments}
This research was supported by the National Key Research and Development Program of China under Grant No.2020AAA0109702, the National Natural Science Foundation of China under Grants 61976208, 62076242, and the InnoHK project.

\bibliography{aaai23}

\end{document}